%% file: main.tex
\pdfoutput=1

\documentclass[11pt]{article}

\usepackage[preprint]{acl}

\usepackage{times}
\usepackage{latexsym}
\usepackage{colortbl}
\usepackage[T1]{fontenc}
\usepackage[utf8]{inputenc}
\usepackage{microtype}
\usepackage{inconsolata}
\usepackage{graphicx}
\usepackage{multirow}
\usepackage{amsmath}
\usepackage{algorithm}
\usepackage{algpseudocode}
\usepackage{tikz}
\usepackage{xspace}

\definecolor{Gray}{gray}{0.9}
\makeatletter
\newcommand{\thickhline}{%
    \noalign {\ifnum 0=`}\fi \hrule height 1.1pt
    \futurelet \reserved@a \@xhline
}

\newcommand{\ALGO}{\texttt{SMITE}\xspace}
\newcommand{\RR}{\ensuremath{\mathcal R}\xspace}
\newcommand{\BB}{\ensuremath{\mathcal B}\xspace}
\renewcommand{\SS}{\ensuremath{\mathcal S}\xspace}
\newcommand{\HH}{\ensuremath{\mathcal H}\xspace}
\newcommand{\PP}{\ensuremath{\mathcal P}\xspace}
\newcommand{\DD}{\ensuremath{\mathcal D}\xspace}
\newcommand{\II}{\ensuremath{\mathcal \lambda}\xspace}

\newcommand{\CC}{\ensuremath{\mathcal C}\xspace}

\newcommand{\PE}{\ensuremath{\mathcal \pi}\xspace}
\newcommand{\FE}{\ensuremath{\mathcal \psi}\xspace}
\newcommand{\EE}{\ensuremath{\mathcal E}\xspace}
\newcommand{\IE}{\ensuremath{\mathcal I}\xspace}
\newcommand{\TE}{\ensuremath{\mathcal T}\xspace}
\newcommand{\DIRatio}{\ensuremath{\mathcal \kappa}\xspace}

\newcommand*\circled[1]{\tikz[baseline=(char.base)]{
            \node[shape=circle,draw,inner sep=2pt] (char) {#1};}}

\title{SMITE: Enhancing Fairness in LLMs through Optimal In-Context Example Selection via Dynamic Validation}

\author{
 \textbf{Garima Chhikara\textsuperscript{1,2}},
 \textbf{Kripabandhu Ghosh\textsuperscript{3}},
 \textbf{Abhijnan Chakraborty\textsuperscript{4}}
\\
 \textsuperscript{1}Indian Institute of Technology Delhi, India \\
 \textsuperscript{2}Delhi Technological University, India \\
   \textsuperscript{3}Indian Institute of Science Education and Research Kolkata, India \\
 \textsuperscript{4}Indian Institute of Technology Kharagpur, India 
}

\begin{document}
\maketitle


\begin{abstract}
Large Language Models (LLMs) are widely used for downstream tasks such as tabular classification, where ensuring fairness in their outputs is critical for inclusivity, equal representation, and responsible AI deployment. This study introduces a novel approach to enhancing LLM performance and fairness through the concept of a dynamic validation set, which evolves alongside the test set, replacing the traditional static validation approach. We also propose an iterative algorithm, \ALGO, to select optimal in-context examples, with each example set validated against its corresponding dynamic validation set. The in-context set with the lowest total error is used as the final demonstration set. Our experiments across four different LLMs show that our proposed techniques significantly improve both predictive accuracy and fairness compared to baseline methods. To our knowledge, this is the first study to apply dynamic validation in the context of in-context learning for LLMs.
\end{abstract}

\input{introduction}
\input{related_work}
\input{methodology}

\input{experiments}

\input{results}

\input{conclusion}

\bibliography{ref}


\end{document}

%% file: introduction.tex
\section{Introduction}
Over the past years, 
Large Language Models (LLMs) have seen rapid growth in their user base and have attracted significant attention from domain experts and the public at large~\cite{rlhf1, NEURIPS2020_1457c0d6, touvron2023llama, openai2023gpt4, geminiteam2023gemini}. In-Context Learning (ICL) capability of LLMs enables them to learn and adapt to a given task solely through the examples provided in the prompt, eliminating the need for further fine-tuning or external training~\cite{NEURIPS2020_1457c0d6, radford2019language}. These examples are referred to as In-Context Examples (ICEs) or In-Context Demonstrations (ICDs). Since the context window for ICL is limited, it is not possible to input overly long texts, making it crucial to strategically select the in-context examples to achieve good performance in a given task
\cite{lu2021fantastically,zhao2021calibrate,li-qiu-2023-finding}.

Some recent works have utilized LLMs for classification of tabular data~\cite{pmlr-v206-hegselmann23a, slack2023tablet, liu-etal-2024-confronting}, where the tabular data is converted into natural language and presented to LLMs along with a brief description of the task to elicit predictions. 
In these tasks, LLMs may inadvertently reinforce social biases embedded in their training data, unless explicitly addressed. This can result in widespread negative consequences, particularly for marginalized and underrepresented groups~\cite{Abid2021, ganguli2022red, hutchinson-etal-2020-social, 10.1145/3531146.3533229, basta-etal-2019-evaluating, askell2021general}. 
Given the growing adoption of LLMs across the software industry, it is crucial to address and mitigate these biases to ensure fair and equitable outcomes.

In this work, we hypothesize that 
intelligent selection of in-context examples can lead to fairer outputs from LLMs. 
Recently, a few works have advocated for selecting a diverse set of in-context examples that is representative of the original dataset, and to check the goodness of the ICE set, typically a validation set is employed which remains the same for all the test cases~\cite{li-qiu-2023-finding, misconfidence}.
In classical machine learning, when the test and validation sets share a similar data distribution, it is easier to obtain better model parameters. However, the test data is not available during model validation. In contrast, with LLMs and in-context learning (ICL), we already have a pre-trained model (the LLM), and the goal is to select the best in-context examples (ICEs) from the available data. For instance, in tabular classification via ICL, we have access to the entire dataset and need to choose examples that can act as in-context demonstrations (ICDs). In ICL, the entire training set can be used to select ICEs for a specific test sample.
We propose a new approach where the validation set used for the selection of examples should closely match the characteristics of the test set. This leads us to introduce the concept of a \textit{dynamic validation set} which adapts based on the test example. Additionally, we propose an iterative greedy algorithm, \ALGO, designed for the selection of examples, that can lead to a strong predictive performance and fairness.
To summarize, our contributions are listed as follows :

\begin{itemize}
    \item We present the idea of employing a \textit{dynamic validation set} (we also call it as \textit{proxy test set}) to measure the performance of current set of in-context demonstrations (Section \ref{sec:proxy_test_examples}).
    \item We propose an iterative algorithm, \ALGO (In-Context Example \textbf{S}election by \textbf{M}inimizing \textbf{I}ndividual and \textbf{T}otal \textbf{E}rror) for the selection of in-context examples. \ALGO calculates the error at each iteration and aims to minimize it in subsequent iterations by selecting elements from the support set (Section \ref{sec:in_context_selection}).
    \item We publicly release the predictions of LLMs for over 1000 test instances across different LLMs, which can spawn future research in this field \footnote{Available at \url{https://anonymous.4open.science/r/ICLSelection-2D3A/}.}.
\end{itemize}

To the best of our knowledge, this work is one of the first efforts to introduce the novel concept of dynamic validation set, wherein the validation set is created by examining the test set. 
Proposed \ALGO is one the earliest algorithms to account for fairness while selecting examples.

%% file: related_work.tex
\section{Related Work}
\label{sec:related_work}

To ensure the reliability of LLMs, it is crucial to eliminate social biases from their outputs. Numerous studies have revealed the existence of bias in the results generated by LLMs~\cite{10.1145/3604915.3608860, bi2023group, Ferrara_2023, nadeem-etal-2021-stereoset, 10.1145/3582269.3615599, bordia-bowman-2019-identifying, freiberger2024fairness, zheng2023large, huang2024bias, chhikara2024fewshotfairnessunveilingllms}.
The output of GPT-3 has shown bias against Muslims \cite{Abid2021}. LLMs demonstrate bias when prompted about stereotypes \cite{10.5555/3666122.3667483}, and the LLMs response exhibit bias even when prompts do not explicitly inquire about it \cite{huang-etal-2021-uncovering-implicit}.

Tabular data is present in many domains \cite{shwartz-ziv2021tabular}. LLMs have been extensively trained on large volumes of natural language data, allowing them to demonstrate remarkable performance across various downstream tasks, even with minimal or no labeled task data \cite{NEURIPS2020_1457c0d6, openai2023gpt4}. Research has proposed methods for inputting tabular data into LLMs to generate predictions \cite{pmlr-v206-hegselmann23a, slack2023tablet}. These predictions must be both fair and accurate, particularly when the tabular data originates from safety-critical and high-stakes domains \cite{9998482, grinsztajn2022why}.

Work by \citealt{liu-etal-2024-confronting} demonstrated that flipping the labels of in-context examples can significantly reduce bias.  
\citealt{bi2023group} introduced a novel chain-of-thought method to diminish biases in LLMs.
Additionally, \citealt{li-qiu-2023-finding} proposed a diversity-guided search which iteratively refines the selected example permutations.
\citealt{misconfidence} focused on selecting demonstrations that minimize the gap between the outputs of LLMs and the actual input-output mappings.


Existing literature has primarily concentrated on the diverse selection of in-context examples and has relied on a static validation set to evaluate the performance of the current set of ICEs. In contrast, we propose a novel concept of a dynamic validation set and an algorithm, \ALGO, designed to select in-context examples that enhance both predictive performance and fairness.


%% file: methodology.tex
\section{Proposed Methodology}

\begin{figure*}[t]
\centering
\includegraphics[scale=0.5,trim={0 0 0 70},clip]{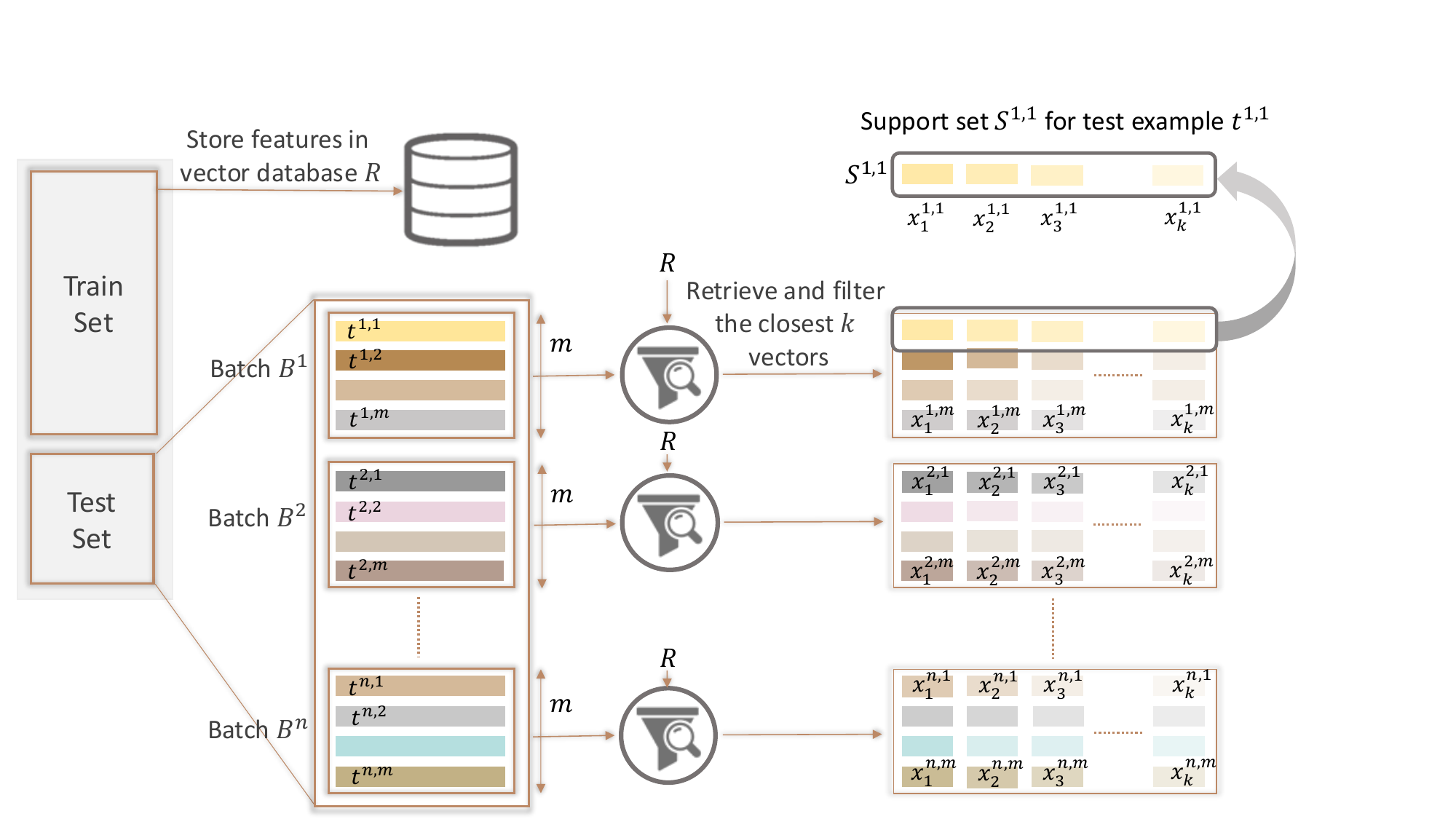}
\caption{Each example in train set is converted to embeddings and stored in a vector database $R$. Test set is divided in $n$ batches each of size $m$. The first test example in $n^{th}$ batch is represented as $t^{n,j}$. For each test example $t^{n,j}$, $k$ closest samples from the vector database $R$ are retrieved and we call it the support set $S^{n,j}$. Support set for a batch $\BB^n$ is defined as the union of the support sets of test examples present in $\BB^n$, i.e. $\SS^n = \bigcup^{m}_{j=1}\{\SS^{n,j}\}$}
\label{fig:support_set}
\end{figure*}

\begin{figure*}[t]
\centering
\includegraphics[scale=0.5,trim={65 320 100 67},clip]{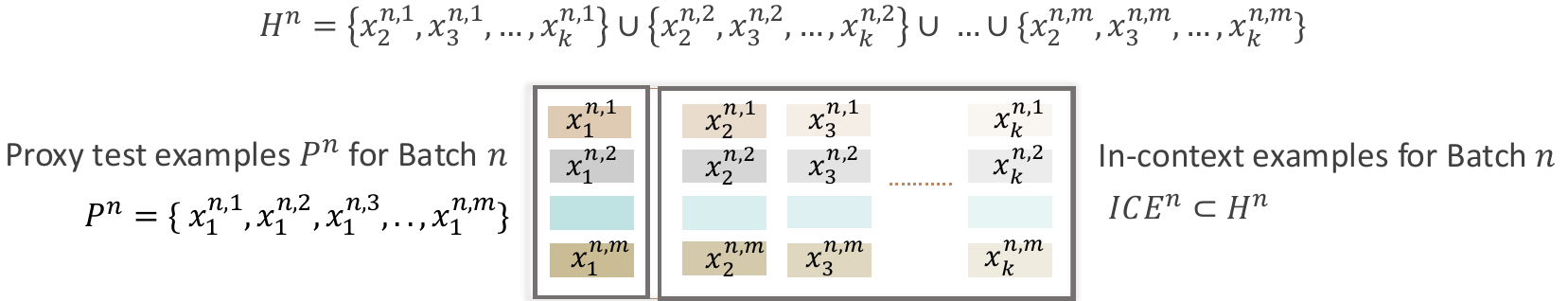}
\caption{Support set $\SS^n$ for a batch $\BB^n$ is divided into two disjoint sets $\PP^n$ and $\HH^n$. $\PP^n$ is the \textit{proxy test set} or the \textit{validation set} which will assess the goodness of the in-context examples. The set $\HH^n$ will be utilized for iteratively selecting the in-context examples.}
\label{fig:vs}
\vspace{-3mm}
\end{figure*}


LLMs are generally used through in-context learning because of the ease of use and easy deployment. 
In ICL our task is to identify the optimal in-context examples that yield better results.
In ICL, we can define two distinct sets: train and test. Our goal is to identify ICEs from the training set while performing inference on the test set. While one might consider using the entire training set as ICEs, the context window size in LLMs is limited to a specific token length, which is insufficient to accommodate the entire dataset. Therefore, careful and strategic selection of ICEs is essential for achieving optimal results from LLMs.

To determine the optimal set of in-context demonstrations for the test examples, we introduce the concept of a \textit{proxy}. As we lack gold standard labels for the test examples, we establish a proxy for these examples, akin to the validation set in classical machine learning (Section \ref{sec:proxy_test_examples}). Subsequently, we present an iterative method, \ALGO, for selecting the ICDs, and utilize the proxy test examples (or dynamic validation set) to assess the quality of these ICDs (Section \ref{sec:in_context_selection}).

\subsection{Selection of Dynamic Validation Set}
\label{sec:proxy_test_examples}

In ICL, we assume a train set $\DD_{train} = \{(x,y,z)\}^{N_{train}}_{i=1}$ and a test set $\DD_{test} = \{(t,y,z)\}^{N_{test}}_{i=1}$ where $x \in \RR$, $t \in \RR$, $y \in \{0,1\}$ denote a binary classification label and $z \in \{0,1\}$ indicate the sensitive features.\footnote{Note that $x$ and $t$ denote the features of the training and testing set respectively, we employ two distinct notations to prevent confusion regarding whether a data point is from the train or the test set.} Labels in the test set are only accessible during evaluations.
$\DD_{train}$ is utilized to find the in-context examples for $\DD_{test}$. 

We convert the examples (also called as the samples or data points) of the training set into embeddings and store it locally in a vector database \textit{R}. Test set $D_{test}$ can be large enough that it may not fit in a single context window, hence we break down the test set in $n$ batches, each of size \textit{m} that is $N_{test} = n*m$. 
Previous studies \cite{li-qiu-2023-finding, misconfidence} have conducted inference on each test example independently. However, a potential drawback of this independent approach is the lack of awareness regarding the sensitive attribute \textit{z} of forthcoming test examples. Our objective is to achieve group fairness, which necessitates being informed about the sensitive attributes of upcoming test examples. Therefore, we process the data in batches rather than making independent decisions for each example, focusing on ensuring fairness within each batch.

The $j^{th}$ example of $i^{th}$ batch is represented as $t^{i,j}$. For each example $t^{i,j}$ we find the closest $k$ vectors from the vector database \textit{R}, such that the sensitive attribute of the test example $t^{i,j}$ is same as the sensitive attribute of the $k$ examples retrieved, and we call it as the \textbf{Support Set} $\SS^{i,j}$. 
$\SS^{i,j} = \{x^{i,j}_1, x^{i,j}_2, ... , x^{i,j}_k\}$, where $x^{i,j}_q$ denotes the datapoint from $D_{train}$ which is $q^{th}$ closest to $t^{i,j}$.
Let $z^{i,j}$ and $z^{i,j}_q$ denote the sensitive attribute for $t^{i,j}$ and and $x^{i,j}_q$ respectively, then $z^{i,j} = z^{i,j}_q \, \forall \, q \in [1,k]$ (Figure \ref{fig:support_set}).

Since we are going to work in batches, let the $n^{th}$ test batch be represented as $\BB^n$ i.e., $\BB^n = \{t^{n,1} , t^{n,2}, ... , t^{n,m}\}$. We define support set $\SS^n$ for $\BB^n$ as $\SS^n = \bigcup^{m}_{j=1}\{\SS^{n,j}\}$    
, i.e., support set of a batch is the union of support sets of each example present in that batch (Figure \ref{fig:vs}). The support set $\SS^n$ is divided into two disjoint sets $\PP^n$ and $\HH^n$ such that $\SS^n = \{\PP^n\} \cup \{\HH^n\}$. $\PP^n$ is a dynamic validation set (or proxy test set) for $\BB^n$, which will validate the selection of in-context examples (discussed in Section \ref{sec:in_context_selection}). We consider the example closest to the test example as its proxy i.e., $\PP^n = \{x^{n,1}_1, x^{n,2}_1, ..., x^{n,m}_1\}$. 
We search for the in-context examples from the set $\HH^n = \{\SS^n\} - \{\PP^n\}$.
One possible approach is to use the entire set $\HH^n$ as in-context demonstrations. However, due to the limited context window size, we must carefully choose the best examples from $\HH^n$, we discuss the selection process in the next section.

\subsection{Selection of In-Context Examples}
\label{sec:in_context_selection}

In this section we discuss about the selection of in-context examples such that both performance and fairness can be optimized, and to validate the performance of in-context selection we utilize the dynamic validation set $\PP^n$ for a given test batch $\BB^n$. 

\subsubsection{Defining the Error Function}
\label{sec:errror}
Let us consider a dynamic validation set $\PP^n = \{x^{n,1}_1, x^{n,2}_1, ..., x^{n,m}_1\}$ and a set of in-context examples $\II$. The gold truth label $y$ for dynamic validation set is known to us, as $\PP^n \subset \DD_{train}$. We also obtain the predictions through the LLMs by passing \circled{a} the instruction about the task, \circled{b} in-context example $\II$, and \circled{c} dynamic validation set $\PP^n$. Performance error \PE is defined as the ratio of incorrect predictions to the total number of predictions made.
\begin{align}
y & = [y^{n,1}_1, y^{n,2}_1, ..., y^{n,m}_1] \\
\hat{y} & = [LLM(Task, \II, \PP^n)] \\
\hat{y} & = [\hat{y}^{n,1}_1, \hat{y}^{n,2}_1, ..., \hat{y}^{n,m}_1] \\
\PE & = 1 - Accuracy(y, \hat{y}) \\
\PE & = \frac{FP + FN}{FP + FN + TP + TN}
\label{eq:performance_error}
\end{align}
Taking motivation from the fairness notion of Demographic Parity \cite{10.1145/3097983.3098095, 10.1145/2783258.2783311, 10.1007/978-3-642-33486-3_3, pmlr-v28-zemel13}, which states that individuals in different groups should have an equal probability of being assigned to the positive predicted class. 
we define fairness error \FE as the absolute difference between the proportion of positive outcomes across sensitive groups. An ideal value of \FE is 0, which indicates that both sensitive groups have been selected proportionally.
\begin{align}
    \FE = abs(P(\hat{y} = 1 | z = 0) - P(\hat{y} = 1 | z = 1))
\label{eq:fairness_error}
\end{align}
We define an error \EE as a combination of performance error \PE and fairness error \FE. The ideal value of \EE is 0, indicating that all predictions are both accurate and fair. The value of $\alpha$ can be determined by the end user based on the desired emphasis is on either performance or fairness (see Algo \ref{algo:error}).
\begin{align}
    \EE = \alpha \cdot \PE + (1-\alpha) \cdot \FE
\label{eq:error}
\end{align}
\begin{algorithm}[t]
\caption{Algorithm for finding the Error \EE}
\label{algo:error}
\begin{algorithmic}
\Function{Error}{$\DD_{train}, \PP^n, \II, \alpha$}
\State $y = \textproc{GetTrueLabel}(\DD_{train}, \PP^n)$
\State $z = \textproc{GetSensitive}(\DD_{train}, \PP^n)$
\State $\hat{y} = \textproc{LLM}(Task, \II, \PP^n)$
\State $\PE = 1 - \textproc{Accuracy}(y, \hat{y})$
\State $p0 = count(\hat{y}=1 \cap z = 0) / count(z = 0)$
\State $p1 = count(\hat{y}=1 \cap z = 1) / count(z = 1)$
\State $\FE = abs(p0-p1)$
\State $\EE = \alpha * \PE + (1-\alpha) * \FE$
\State return $\EE$
\EndFunction
\end{algorithmic}
\end{algorithm}

\begin{figure*}[t]
\centering
\includegraphics[scale=0.5,trim={20 20 20 30},clip]{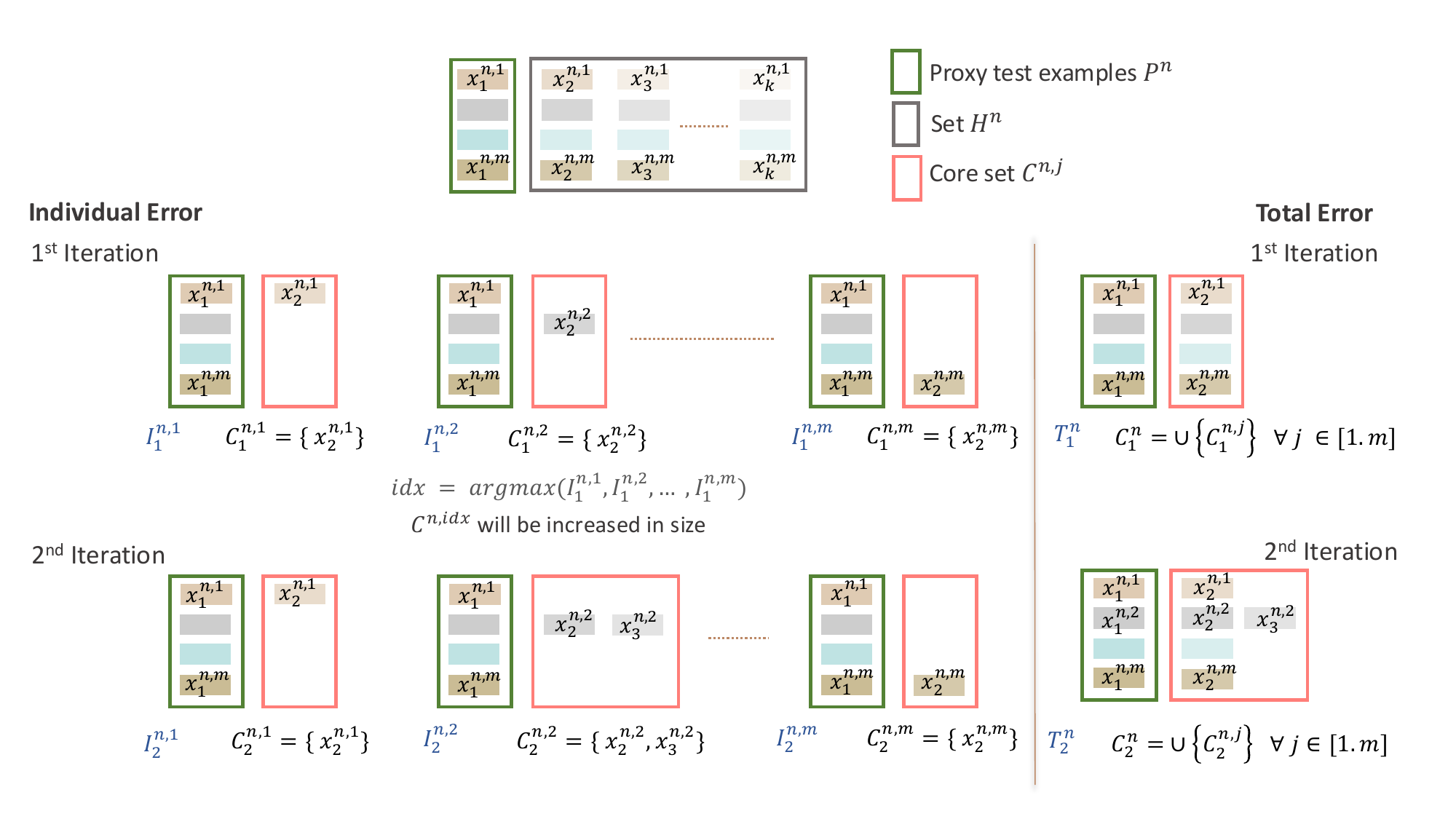}
\caption{Corresponding to each test example $t^{n,j}$ we specify a core set $C^{n,j}$. $C^{n,j}$ is initialized with the second closest example to $t^{n,j}$. Effectiveness of each core set is calculated using individual error \IE. Core set which yields the the maximum error \IE, requires an extra support, as it in itself is not able to give good results, hence we increase the size of core set by adding the next closest example. $C^{n,j}_r$ denotes the core set $C^{n,j}$ in the $r^{th}$ iteration. $\IE^{n,j}_r$ denotes the individual error w.r.t core set $C^{n,j}$ in the $r^{th}$ iteration. In each iteration we also calculate the total error \TE using $C^{n}_r$ where $\CC^n_r = \bigcup^{m}_{j=1}\{\CC^{n,j_r}\}$. $\CC^n_r$ corresponding to the minimum error \TE is selected as the final in-context demonstration set.}
\label{fig:iptp}
\vspace{-3mm}
\end{figure*}

\subsection{\ALGO for ICD Selection}
In this section, we discuss about our proposed algorithm \ALGO (In-Context Example \textbf{S}election by \textbf{M}inimizing \textbf{I}ndividual and \textbf{T}otal \textbf{E}rror).
Dynamic validation set is the one which is closest to the test example. For a given test batch $\BB^n$, our aim is to find the best set of in-context examples from set $\HH^n$.

Corresponding to each test example $t^{n,j}$ we define a core set $\CC^{n,j}$. Set $\CC^{n,j} = \{x^{n,j}_2\} \; \forall \; j \in [1,m]$, i.e., the set $\CC^{n,j}$ is initialized with the second closest example to $t^{n,j}$.
Calculate the error with respect to each $\CC^{n,j}$ by utilising $\CC^{n,j}$ as the in-context example \II. We call this error as the \textit{individual error} \IE, name \textit{individual} because here we check the effectiveness with regard to each set $\CC^{n,j}$.
Calculate $\IE^{n,j} \; \forall \; j \in [1,m]$.
Idea behind this is to know which $\CC^{n,j}$ is the least effective and requires more \textit{support} from the support set. 
The core set $\CC^{n,j}$ corresponding to the maximum $\IE^{n,j}$ will be increased 
in size by adding the next closest example from the support set $\SS^{n,j}$ (refer Figure \ref{fig:iptp}). 

We define another set $\CC^n = \bigcup^{m}_{j=1}\{\CC^{n,j}\}$, which is the union of all the core sets. We calculate the error by utilizing $\CC^n$ as the in-context example \II, and name this as \textit{total error} \TE. Total error corresponding to $\CC^n$ is $\TE^n$.

Iteratively repeat this process of computing \IE and \TE by increasing the core set, $l$ number of times, where $l$ is a hyper-parameter.
The set $\CC^n$ corresponding to the minimum $\TE^n$ is chosen as the final in-context demonstration set for batch $\BB^n$ (refer Algo \ref{algo:icl}). 

\begin{algorithm}[t]
\caption{Algorithm for finding the in-context examples for a batch}
\label{algo:icl}
\begin{algorithmic}
\Function{ICDSelect}{${\DD_{train}, \PP^n, \HH^n, \alpha, l, m}$}

\State $li = [\;]$ \Comment{$li$ denotes the last index}
\For{$j \gets 1$ to $m$}
    \State $\CC^{n,j} = \{x^{n,j}_2\}$ \Comment{Initialize $\CC^{n,j}$}
    \State $li.append(2)$
\EndFor

\State $icd = [\;]$ \Comment{Stores the final result}
\State $min = 1$ \Comment{Keeps track of minimum error}

\For{$r \gets 1$ to $l$}

    \For{$j \gets 1$ to $m$}
        \State $\IE^{n,j} = \textproc{Error}(\DD_{train}, \PP^n, \CC^{n,j}, \alpha)$
    \EndFor

    \State $\CC^{n} = \{\}$
    \For{$j \gets 1$ to $m$}
        \State $\CC^{n} = \CC^{n} \cup \CC^{n,j}$
    \EndFor

    \State $\TE^{n} = \textproc{Error}(\DD_{train}, \PP^n, \CC^{n}, \alpha)$

    \If{$\TE^n < min$}
        \State $min = \TE^n$
        \State $icd = li.copy()$
    \EndIf
    \State $idx = argmax(\IE^{n,1}, \IE^{n,2}, ... , \IE^{n,m})$
    \State $li[idx] = li[idx] + 1$
    \State $\CC^{n,idx} = \CC^{n,idx} \cup x^{n, idx}_{li[idx]}$
\EndFor
\State return $icd$
\EndFunction
\end{algorithmic}
\end{algorithm}

%% file: experiments.tex
\section{Experimental Setup}
In this section, we outline the overall setup of the experiments, covering aspects such as the dataset, models utilized, different performance and fairness metrics used for evaluation.

\subsection{Dataset}
\label{sec:dataset}
We utilize two popular tabular datasets -- \textit{UCI Adult Income} \cite{misc_adult_2} and \textit{COMPAS} Dataset \cite{compas}, which have been extensively used to study about fairness in machine learning. 

The \textit{Adult} \cite{misc_adult_2} dataset is derived from the 1994 U.S. Census Bureau database. The objective is to predict whether an individual earns >\$50K or <=\$50K per year based on the profile data. 
We refine the \textit{Adult} dataset by removing the dependent features, such as \textit{education} and \textit{education-num}, both denotes the level of education, we only select the attributes which are independent and cannot be derived from other attributes. We select 12 features to conduct the experiments - \textit{age}, \textit{workclass}, \textit{education}, \textit{marital status}, \textit{occupation}, \textit{relationship}, \textit{race}, \textit{sex}, \textit{capital gain}, \textit{capital loss}, \textit{hours per week} and \textit{native country}. The target variable \textit{income} takes on a binary value, either <=50K or >50K.
We refine the dataset by removing rows that contain null or irrelevant values, resulting in a final dataset of 44,869 rows each comprising 12 features. We use 1,000 samples in the test set and rest 43,869 examples for the train set.

The \textit{COMPAS} \cite{compas} dataset was collected as part of the ProPublica analysis of machine bias in criminal sentencing. Target is to predict whether a person will recidivate or not. We utilize 10 features to make the predictions - \textit{age}, \textit{charge degree}, \textit{race}, \textit{age}, \textit{category}, \textit{score text}, \textit{sex}, \textit{priors count}, \textit{decile score} and \textit{days inside jail}. The target variable \textit{two year recidivate} takes binary value i.e., either 0 or 1; 0 indicating person will not recidivate and 1 indicating person will recidivate. 
After cleaning the dataset we obtain 5,278 samples, out of which we utilize 1,000 for test set and other 4,278 for train set. 

Our experimentation on \textit{Adult} and \textit{COMPAS} dataset primarily focuses on \textit{sex} as the protected attribute.

\subsection{Large Language Models}
LLMs are characterized by their extensive parameter sizes and remarkable learning abilities \cite{zhao2023survey, chang2023survey}. 
In our work, we utilize three open-source LLMs and one proprietary LLM to conduct experiments: \texttt{llama3-70b-8192} from Meta \cite{touvron2023llama}, \texttt{mixtral-8x7b-32768} from Mistral AI \cite{jiang2024mixtral}, \texttt{gemini-1.0-pro} from Google \cite{geminiteam2023gemini} and \texttt{gpt-4o-mini-2024-07-18} from OpenAI \cite{gpt4omini}.
Across all experiments, we keep temperature, top probability and output tokens as 0, 0.9 and 8192 respectively.

\begin{table*}[t]
\centering
\begin{tabular}{l | c  c  c  c  c | c  c | c} 
\thickhline
\multirow{2}{*}{Models} & 
\multicolumn{5}{c}{Prediction Metrics} & 
\multicolumn{2}{|c|}{Fairness Metrics}  &
\multirow{2}{*}{Score \EE $\downarrow$} \\
\cline{2-8}
& Accuracy $\uparrow$ & Precision $\uparrow$  & Recall $\uparrow$ & F1 Score $\uparrow$ & \PE $\downarrow$ & \FE $\downarrow$ & \DIRatio $\downarrow$& \\
\thickhline
\rowcolor{Gray} \multicolumn{9}{c}{Zero Shot} \\
\thickhline
Llama  & 0.728&	0.739&	0.728&	0.725 &0.272 &0.228	&0.359 &0.250  \\
Mixtral & 0.634	&0.686	&0.634	&0.602& 0.366 & 0.212 &	0.632 & 0.289\\
Gemini & 0.670	& 0.681	& 0.670	&0.666 &0.330 &0.340	&0.519 & 0.335\\
GPT & 0.711	&0.731	&0.711	&0.705& 0.289 &0.302&	0.644 &0.296\\
\thickhline

\rowcolor{Gray} \multicolumn{9}{c}{Few Shot Random ICE} \\
\thickhline
Llama & 0.725&	0.740	&0.725	&0.720	&0.275 &0.218	&0.399 & 0.246 \\
Mixtral  & 0.736&	0.749	&0.736	&0.732	& 0.264 &0.228	&0.406	& 0.246 \\
Gemini & 0.698	&0.711	&0.698	&0.694 &0.302& 0.352	&0.509 &0.327\\
GPT & 0.740	&0.761&	0.740	&0.735 &0.260 &0.172	&0.409& 0.216\\
\thickhline

\rowcolor{Gray} \multicolumn{9}{c}{Few Shot RAG} \\
\thickhline
Llama & 0.774	&0.794	&0.774&	0.770 &0.226 &0.184	&0.395 &0.205  \\
Mixtral & 0.755	&0.785	&0.755&	0.747	&0.245&0.186	&0.491	&0.215 \\
Gemini & 0.719	&0.743	&0.719	&0.712 &0.281 &0.230 &	0.532 &0.256\\
GPT & 0.735&	0.762&	0.735	&0.726 &0.265 &0.162	&0.490 &0.213\\
\thickhline


\rowcolor{Gray} \multicolumn{9}{c}{Proposed \ALGO Algorithm} \\
\thickhline
Llama & 0.758	&0.786	&0.758	&0.75 & 0.242 & 0.148 & 0.404 & \textbf{0.148} \\
Mixtral  & 0.746 & 0.779	&0.746	&0.736 &0.254 &0.150	&0.380& \textbf{0.202} \\
Gemini & 0.739	&0.755	&0.739	&0.734 &0.261 &0.202	&0.416 &\textbf{0.232}\\
GPT &0.740	&0.780&	0.740	&0.730 &0.260 &0.133	&0.373 & \textbf{0.197}\\
\thickhline

\end{tabular}
\caption{Results obtained using various baselines and our proposed \ALGO algorithm for \textbf{\textit{Adult}} Dataset. Error \EE is calculated as $\EE = \alpha * \PE + (1-\alpha) * \FE$ where value of $\alpha$ is set to 0.5. Selection of in-context examples through \ALGO yields the best results across all LLMs. It is important to note that values close to 1 are preferred for Accuracy, Precision, Recall, and F1-Score, while lower values for \PE, \FE, \DIRatio, and \EE (ideally 0) indicate better outcomes.}
\label{tab:adult_result}
\end{table*}

\begin{table*}[t]
\centering
\begin{tabular}{l | c  c  c  c  c | c  c | c} 
\thickhline
\multirow{2}{*}{Models} & 
\multicolumn{5}{c}{Prediction Metrics} & 
\multicolumn{2}{|c|}{Fairness Metrics}  &
\multirow{2}{*}{Score \EE} \\
\cline{2-8}
& Accuracy $\uparrow$ & Precision $\uparrow$  & Recall $\uparrow$ & F1 Score $\uparrow$ & \PE $\downarrow$ & \FE $\downarrow$ & \DIRatio $\downarrow$& \\
\thickhline
Llama & 0.645&	0.656&	0.645	&0.638& 0.355 &0.218	&0.721 & 0.287\\
Mixtral & 0.614	&0.629	&0.614	&0.603 &0.386 &0.260	& 0.791 & 0.323 \\
Gemini & 0.612	&0.623	&0.612	&0.603& 0.388 &0.400&	0.540 & 0.394\\
GPT & 0.616	&0.631	&0.616	&0.607 &0.384 &0.320	&1.042& 0.352\\
\thickhline

\rowcolor{Gray} \multicolumn{9}{c}{Few Shot Random ICE} \\
\thickhline
Llama & 0.642&	0.659&	0.642&	0.632 &0.358 &0.232	&0.694 & 0.295 \\
Mixtral  & 0.623	&0.633	&0.623	&0.615 &0.377 &0.226	&0.596 & 0.302  \\
Gemini & 0.620	&0.647	&0.620	&0.603 &0.380& 0.216&	0.481 &0.298\\
GPT & 0.632	&0.654&	0.632	&0.616& 0.368 &0.216&	0.400 &  0.292\\
\thickhline

\rowcolor{Gray} \multicolumn{9}{c}{Few Shot RAG} \\
\thickhline
Llama & 0.644	&0.653	&0.644	&0.639 &0.356 &0.212	&0.656 &0.284 \\
Mixtral & 0.637	&0.646	&0.637&	0.632& 0.363 &0.238&	0.871 & 0.300 \\
Gemini & 0.636&	0.645&	0.636	&0.63 &0.364 &0.236&	0.667 & 0.300 \\
GPT & 0.644 &	0.655	&0.644	&0.638 &0.356 &0.196	&0.571 & 0.276 \\
\thickhline


\rowcolor{Gray} \multicolumn{9}{c}{Proposed \ALGO Algorithm} \\
\thickhline
Llama & 0.646 &	0.666	&0.646	&0.635 &0.354 &0.172	&0.390 &\textbf{0.263}  \\
Mixtral & 0.648	&0.662	&0.648	&0.642& 0.352  &0.208	&0.630 &\textbf{0.280}\\
Gemini & 0.647	&0.657	&0.647	&0.637 &0.353 &0.190	&0.652 &\textbf{0.271}\\
GPT & 0.662	 & 0.68 &	0.662	& 0.651  & 0.338  & 0.172&	0.356& \textbf{0.255} \\
\thickhline

\end{tabular}
\caption{Results obtained using various baselines and our proposed \ALGO algorithm for \textbf{\textit{COMPAS}} Dataset. Selection of in-context examples through \ALGO yields the best results across all LLMs.}
\label{tab:compas_result}
\end{table*}

\subsection{Metrics}
We evaluate the prediction performance of the LLMs using metrics such as Accuracy, Precision, Recall, F1-Score and performance error \PE (Equation \ref{eq:performance_error}). 
We utilize two variations of disparate impact as the fairness metrics.
Disparate impact \cite{10.1145/2783258.2783311} assesses the probability of being positively classified  (Equation \ref{eq:disparate_impact}). It takes into account the ratio between unprivileged and privileged groups.
\begin{equation}
    DI = \frac{P(\hat{y} = 1 | z = 0)}{P(\hat{y} = 1 | z = 1)} =
    \frac{\frac{TP_0 + FP_0}{N_0}}{\frac{TP_1 + FP_1}{N_1}}
\label{eq:disparate_impact}
\end{equation}
A value close to 1 is considered ideal for $DI$, as it signifies an equitable distribution across both demographic groups. For our experiments, we utilize \FE (Equation \ref{eq:fairness_error}) and \DIRatio as the fairness metrics. 
\begin{align}
\DIRatio & = abs(1 - DI) \\
\DIRatio &= abs \left( 1 - \frac{P(\hat{y} = 1 | z = 0)}{P(\hat{y} = 1 | z = 1) + \varrho} \right) \\ 
\DIRatio &= abs \left(1- 
    \frac{\frac{TP_0 + FP_0}{N_0}}{\frac{TP_1 + FP_1}{N_1} + \varrho} \right)
\label{eq:diratio_error}
\end{align}
$\varrho$ is added in the denominator to avoid divide by zero error, $\varrho$ is set to a very small value such as $10^{-5}$. Ideal value of \DIRatio is 0, where 0 signifies an equal distribution across demographics, ensuring fairer results. 

\subsection{Input Parameters}
For both \textit{Adult} and \textit{COMPAS} dataset, number of examples in test set are 1000, i.e., $N_{test} = 1000$. For selection of test set we experimented with three different seed values [20, 25, 42] and for each seed we ran the experiments three times to ensure the reliability of results. Value of $n$, $m$, $k$ are set to 50, 20 and 15 respectively. Iterative algorithm \ALGO is executed $l$ times, $l$ is set as 10. For all the experiments value of $\alpha$ was set to 0.5, ensuring equal importance to both performance and fairness. 

We utilize the LangChain\footnote{\url{https://python.langchain.com/docs/tutorials/rag/}} framework of Retrieval Augmented Generation (RAG), to store the data into vector database and to retrieve the relevant data.  
OpenAI Embeddings\footnote{\url{https://platform.openai.com/docs/guides/embeddings/embedding-models}} are utilized to convert the examples of training set from text to vectors, and are then stored in Chroma\footnote{\url{https://python.langchain.com/docs/integrations/vectorstores/chroma/}} vector database.

%% file: results.tex
\section{Results}

\subsection{Comparison with the Baseline}
We employ three different baselines to evaluate the results obtained through \ALGO. i). Zero Shot Learning where we do not provide in context examples, prompt constitutes of the task instruction and the test example. ii). Few Shot with Random selection of ICE, here we do a random selection of in-context examples. iii). Few Shot RAG where we select the most closest example to the test set as the in-context demonstrations. For each test instance in the batch we select the closest example. Essentially, in this case, the dynamic validation set acts as in-context demonstrations, and it is a single step task instead of an iterative procedure.

Our results in Table \ref{tab:adult_result} and \ref{tab:compas_result} indicate that \ALGO performs the best across all the baselines in terms of both performance error \PE and fairness error \FE.
\EE defines a linear combination of \PE and \FE, it can be observed that for all the LLMs error \EE reduces when in-context demonstrations are selected using \ALGO. 
In \textit{Adult} dataset, Llama achieved the best result with the error value being the minimum as 0.148, followed by GPT, Mixtral and Gemini. A notable difference is observed through the usage of our proposed \ALGO algorithm. 
Llama, for example, had an error of 0.246 when using random in-context examples, but this was reduced to 0.148 with \ALGO, highlighting the effectiveness of our proposed approach.
For \textit{COMPAS} dataset, GPT model gives the best performance, followed by Llama, Gemini and Mixtral. 


\subsection{Accuracy-Fairness Trade off}
Generally there is a trade off between accuracy and fairness. As we try to improve the fairness, accuracy generally drops. This trade off can be controlled through the parameter $\alpha$ in Equation \ref{eq:error}. Throughout our experiments we keep value of $\alpha$ as 0.5, emphasizing on both the performance and fairness. 
\ALGO gives better or similar results in terms of predictive performance when compared to baseline, while at the same time improving the fairness results. 


%% file: conclusion.tex
\section{Concluding Discussion}
This work proposes a novel concept of \textit{Dynamic Validation Set} or \textit{Proxy Test Set}, which can be used to validate the performance of in-context examples. In classical ML, we try to keep the distribution of validation set and test set similar, motivated by the same idea we proposed a method which selects the validation set which is close to the test set for ICL. 
Further, we propose an algorithm \ALGO which demonstrates a greedy iterative strategy for selection of ICDs by the usage of Individual Error \IE and Total Error \TE. Experimental results revealed that usage of proxy test examples and \ALGO can achieve better performance and fairness as compared to the baselines. 

\subsection*{Limitation}
Our proposed \ALGO is designed to handle the criteria of group fairness. It may not work for individual fairness, some modifications might be required to adapt to other fairness notion. Our focus was on binary classification task with a single sensitive attribute, which might not always align with the practical situation. Dynamic validation set is selected at the inference time, hence the proposed method \ALGO can have a higher run-time. 